\documentclass{article} 
\usepackage{MLDD_workshop_2022,times}

\usepackage{amsmath,amsfonts,bm}

\def\eqref#1{equation~\ref{#1}}

\def\1{\bm{1}}

\DeclareMathAlphabet{\mathsfit}{\encodingdefault}{\sfdefault}{m}{sl}
\SetMathAlphabet{\mathsfit}{bold}{\encodingdefault}{\sfdefault}{bx}{n}

\usepackage{hyperref}
\usepackage{url}
\usepackage{graphicx}

\title{Partial Product Aware Machine Learning on DNA-Encoded Libraries}

\author{Polina Binder, Meghan Lawler, LaShadric Grady, Neil Carlson,  Sumudu Leelananda,\\
\textbf{Svetlana Belyanskaya, Joe Franklin, Nicolas Tilmans, Henri Palacci} \\
Anagenex \\
San Francisco, CA  94117, USA \\
\texttt{\{polina,meghan,lashadric,neil,sumudu} \\
\texttt{svetlana,joe,nicolas,henri\}@anagenex.com} \\
}

\iclrfinalcopy 
\begin{document}

\maketitle

\begin{abstract}
DNA encoded libraries (DELs) are used for rapid large-scale screening of small molecules against a protein target. These combinatorial libraries are built through several cycles of chemistry and DNA ligation, producing large sets of DNA-tagged molecules. Training machine learning models on DEL data has been shown to be effective at predicting molecules of interest dissimilar from those in the original DEL. Machine learning chemical property prediction approaches rely on the assumption that the property of interest is linked to a single chemical structure. In the context of DNA-encoded libraries, this is equivalent to assuming that every chemical reaction fully yields the desired product. However, in practice, multi-step chemical synthesis sometimes generates partial molecules. Each unique DNA tag in a DEL therefore corresponds to a set of possible molecules. Here, we leverage reaction yield data to enumerate the set of possible molecules corresponding to a given DNA tag. This paper demonstrates that training a custom GNN on this richer dataset improves accuracy and generalization performance.
\end{abstract}

\section{Introduction}

Discovery of new small molecules that can be used as pharmaceutical drugs often starts with screening against a target protein of interest involved in a disease pathway.
DNA Encoded Libraries (DELs) are used to rapidly screen large, diverse sets of small molecules against such proteins \cite{clark2009design}. In these libraries, individual molecules are constructed from simpler molecular fragments called building blocks that are combined through several cycles of chemistry and DNA ligation, producing a large number of DNA-barcoded molecules. Libraries are produced through a massively parallelized  method, “split-and-pool” where after each cycle of chemistry and DNA  ligation, the products are pooled into one well and then dispersed out for the next round of chemistry and DNA ligation \cite{clark2010selecting}.  This allows for the production of 
libraries which contain over $10^8$ unique DEL molecules. These molecules are incubated with the target protein, and remaining molecules that are not bound are washed away, so only the bound molecules remain. The bound molecules are disassociated from the target and the process is repeated, resulting in an increased proportion of molecules that bind the target remaining after each round of selection. The DNA is then amplified and sequenced to identify the molecules that bind to the protein of interest. 
\begin{figure}[h]
\begin{center}
\includegraphics[scale = 0.18]{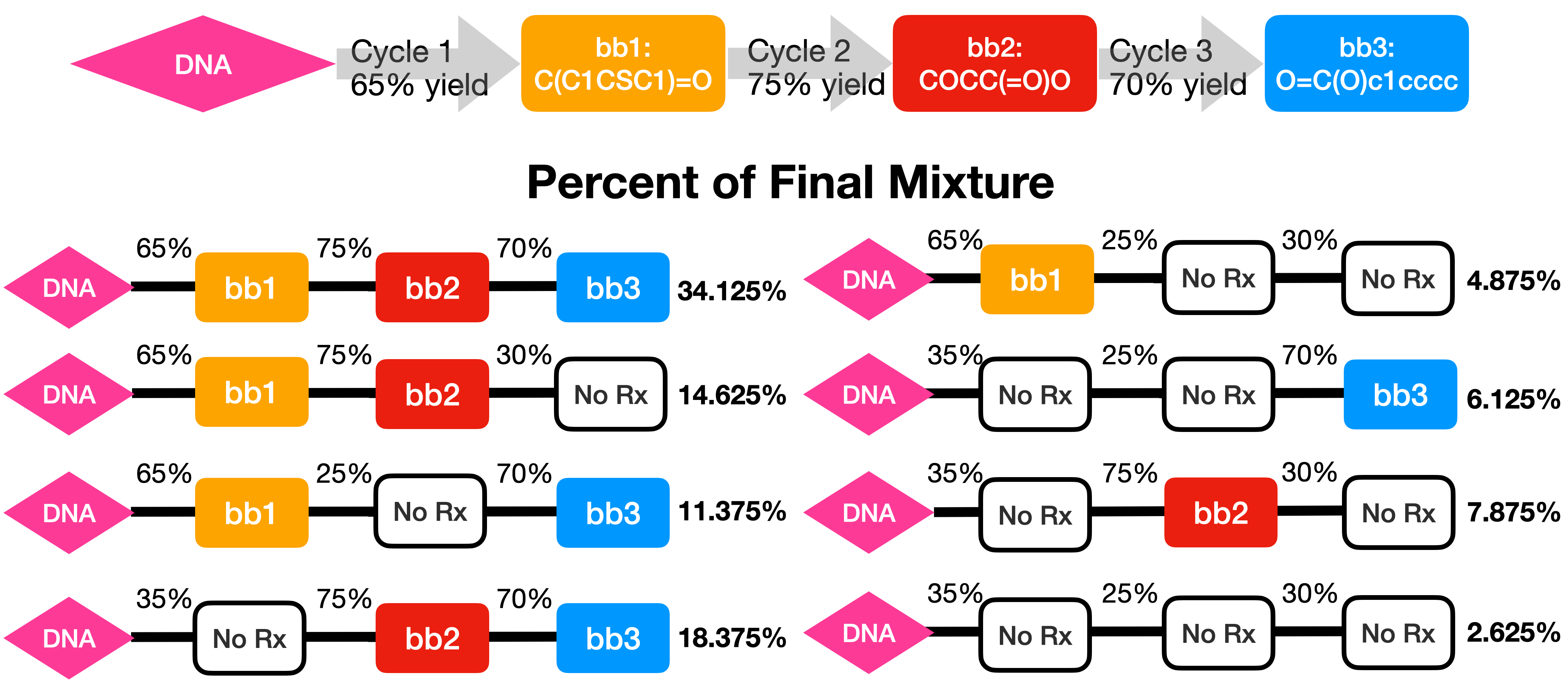}

\end{center}
\caption{Mixture of molecules in a three cycle library. Simulated yields and corresponding proportions of products in the final mixture are shown.}\label{mixtures}
\end{figure}

The DEL synthesis process can, however, cause several chemical structures to carry the same DNA barcode. When a building block is added at each cycle, a chemical reaction occurs and only a percentage of the reaction's products produce the desired output.  This percentage is referred to as the reaction yield. As the DNA tag is added at each cycle regardless of the completion
of the chemistry reaction, a full DNA barcode may correspond instead to an incomplete (partial) product or an unexpected byproduct \cite{satz_simulated_2016}. For a three cycle library and a given DNA barcode, there are three reaction steps each of which succeeds with a given probability. Hence, a given barcode may correspond to a three cycle product (trisynthon), a two cycle product (disynthon), a one cycle product (monosynthon), or other possible side products. We can estimate the reaction yields before library construction by quantifying individual building block yields in a set of validation experiments. The possible products for a three cycle library (excluding byproducts) are shown in figure \ref{mixtures} along with sample proportions in the final mixture.

Multiple libraries of DEL molecules can be pooled together. This pool of molecules is incubated with the target protein, unbound molecules are washed away and bound molecules are separated from the target protein. This process is repeated three times to eliminate non-binding molecules \cite{clark2009design}. The remaining molecules' DNA tags can then be amplified and sequenced, thus identifying potential binders.

A source of variability in the sequencing read count is that the molecules will not have uniform counts in the starting pool. To account for this effect, the pool can be sequenced before selection (dilute library sequencing or DLS) \cite{variableselect}.

Additionally, molecules could come through in sequencing simply because they bind the matrix and not the target protein itself. To measure this effect, the pools can also be screened without the presence of the target. The counts from this experiment are referred to as the "no target control" (NTC) counts.  Some molecules could also appear to be enriched in selections for several targets, indicating that the molecule may be promiscuous and bind to several targets indiscriminately \cite{satz_simulated_2016}.

In this paper, we formulate a model that incorporates partial products in the probabilistic modeling of the counts, using the probabilistic framework described in  \cite{lim_machine_2021}. We use a single GNN to predict enrichments of the full product, but also to predict enrichments of partial products.  These enrichments are then combined to estimate the read counts for a given DNA sequence. To estimate the weights of the partial products, we show that using the yields from the validation experiments provide a marked improvement over using constant data. 

To evaluate the generalization capabilities of our model trained on DEL data, we examine the performance of our model on molecules from external vendors that are different from those in the synthesized libraries. In this work, we use a validation set of 142 molecules  from an external vendor that were screened against the protein of interest. The binding affinities were determined based on experiments performed in our own laboratory, producing a consistent, comparable training set. 

\section{Related Work}

Several approaches to training machine learning models on DNA-encoded library selection data have been recently been proposed. \cite{mccloskey_machine_2020} de-noised three cycle DEL data by aggregating counts across all combinations of two building blocks and training a GNN on the resulting dataset. This approach can be interpreted as a way to only model 2-cycle partial products with constant proportions, without taking into account the full three-cycle product. \cite{lim_machine_2021} presented a negative log-likelihood regression modeling approach to DELs that took into account both the binding counts to the target and to the NTC on trisynthons. Finally, \cite{ma_regression_2021} presented a negative binomial regression modelling approach trained on trisynthon data. Their approach provides a principled way to incorporate all the factors driving read counts, such as matrix binding, promiscuity, target binding. We leverage this flexible framework to incorporate partial product information.

The neural network architecture employed in this paper is a message passing graph neural network \cite{gilmer2017neural}. GNNs have been successfully used for molecular property prediction in DELs \cite{lim_machine_2021,ma_regression_2021,mccloskey_machine_2020} and also in a variety of property prediction problem formulations \cite{chen2018rise}.

Work by \cite{komar_denoising_2020} presented an approach for de-noising DNA encoded libraries that included modeling the yield data and modeling partial products but did not provide a model for predicting affinity on new molecules.

\section{Methodology}
\subsection{Dataset}
To build the dataset, we use a combination of fifteen proprietary DEL libraries, totalling 700 million unique molecules. These libraries were built in-house and screened in-house against a difficult nucleic acid binding protein target. The resulting selection was then sequenced, producing 200 million individual sequence reads corresponding to 90 milllion individual molecules. From these, about 4 million molecules were determined to be significant binders to the protein target. 

To increase the number of negative examples, we added one million molecules that were sequenced but bound only to the NTC or to other targets and another million molecules that were not sequenced but were present in the initial libraries. We split this dataset into train and test sets using a Murcko scaffold split \cite{rdkit}. 

\subsection{Partial product data}

To enrich this dataset with partial products, we computed all the possible two-building block and three-building block molecules corresponding to a given DNA sequence. Initial tests including the single building block partial products did not lead to any improvement in model performance and were therefore excluded from the final dataset. The theoretical proportion of each partial product was then computed from the estimated building block yields.
\begin{figure}[h]
\begin{center}
\includegraphics[scale = 0.22]{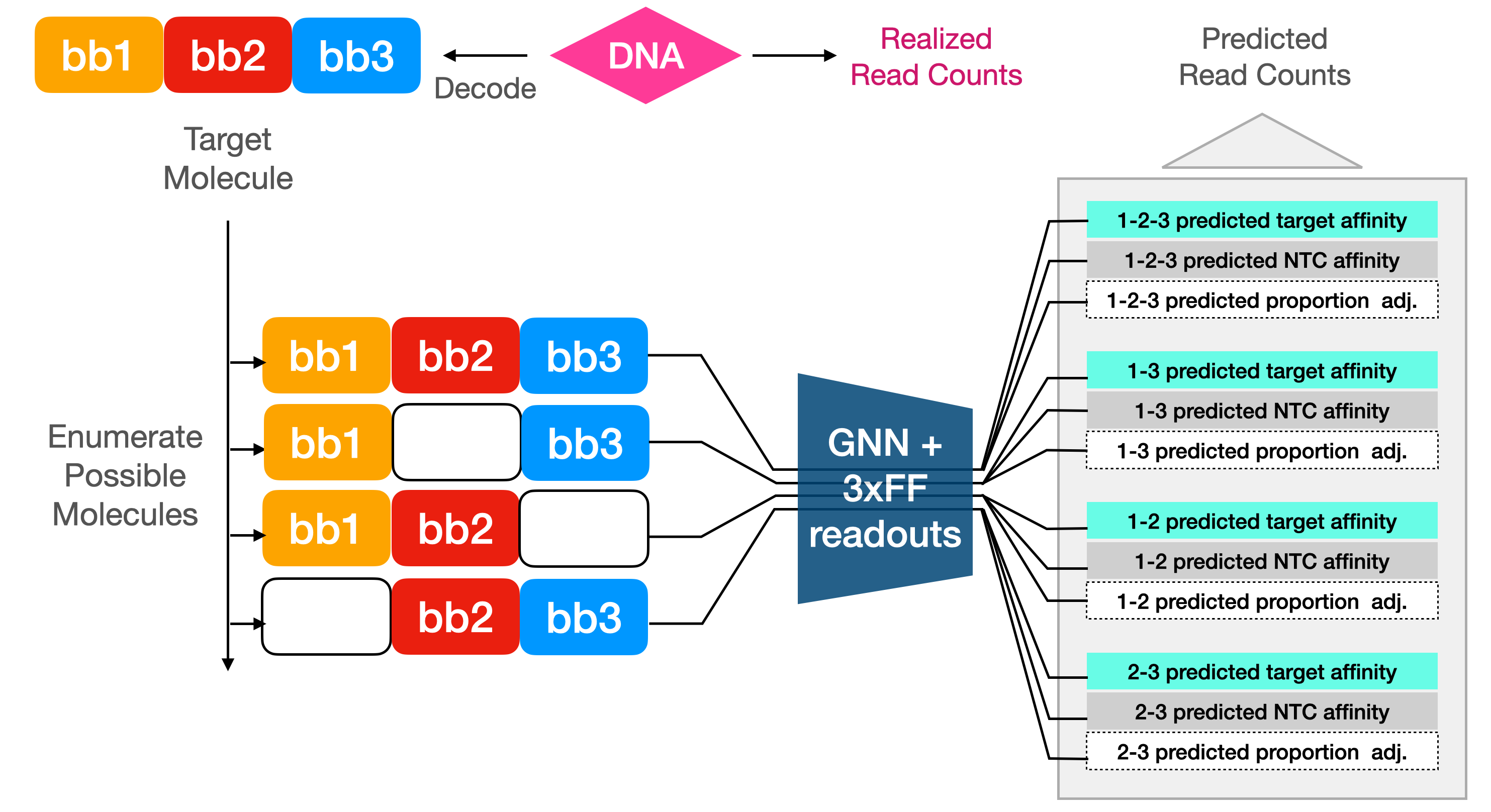}

\end{center}
\caption{An overview of the model architecture. The trisynthon and disynthons are enumerated. Each of them is passed through a GNN which outputs the predicted target enrichment (affinity), predicted NTC enrichment, and the predicted adjustment to the product proportion. These enrichments can be directly used in the testing phase, or combined with other factors to obtain the predicted counts.}
\label{model}
\end{figure}
\subsection{Model}

The goal of this work is to learn a model that outputs the enrichment of a molecule $i$ against a target, $R_{target, i}$, and against the NTC, $R_{NTC, i}$. This model can then be used to screen commercial libraries for other molecules that bind to the target of interest.

Following \cite{ma_regression_2021}, we model the read counts as following a negative binomial distribution whose mean is driven by multiple factors.   However, for a given DNA tag/read count pair, we take into account both full and possible partial products as being factors driving the total read count. An overview of the approach and model architecture is shown in figure \ref{model}.

The read count of a DNA tag $i$ in either the NTC or target selection experiment is modeled by a negative binomial distribution with a mean parameter $\mu_i$ and a dispersion parameter $\alpha: C_{target,i}\sim \textit{NB}(\mu_{target,i}, \alpha_{target})$ and $C_{NTC,i}\sim \textit{NB}(\mu_{NTC,i}, \alpha_{NTC})$. The $\alpha$ values are obtained through a negative binomial regression before training. 
\newpage
 
For the $\mu_i$ values, we assume that binding enrichment of the molecules corresponding to tag $i$ against a given target ($B_{target, i}$) is due to a combination of the enrichment of the corresponding trisynthon ($R_{tri,i}$) and three possible disynthons ($R_{di_1,i}$,$R_{di_2,i}$, $R_{di_3,i}$):

\begin{equation}
B_{target, i}=p_{tri, i}R_{tri, i} + \Sigma_{j=1}^{3}p_{di_j,i}R_{di_j,i}
\end{equation}

where the $p$ values correspond to the proportions of the trisynthons and disynthons that are present in the final mixture.
Similarly, we assume that enrichment of the molecules in the NTC ($B_{NTC, i}$)  is also due to a combination of the enrichment in the trisynthon and disynthons: 

\begin{equation}
B_{NTC,i}=p_{ tri,i}R_{NTC,tri,i} + \Sigma_{j=1}^{3}p_{di_j,i}R_{NTC,di_j,i}.
\end{equation}

We assume that the counts for a given DNA tag against the target are due to a combination of the binding enrichment to the target, the binding enrichment in the NTC experiment, the starting material estimated using the DLS experiment ($C_{dls,i}$), and promiscuous binding counts ($C_{promiscuity, i}$). 

Thus, we set the mean parameter of the negative binomial distribution as 
\begin{equation}
\mu_{i,target} = \sigma(B_{target,i} + \beta_{NTC}\cdot B_{NTC,i} + \beta_{DLS} p_{dls_i}+\beta_{promiscuity}c_{promiscuity} +\beta_{constant})
\end{equation}

Here, the $\beta$ values are learned by negative binomial regression on the training set and $\sigma$ is the softplus function.  
Similarly, the counts for a given DNA tag against the NTC are explained by the binding enrichment to the NTC, the DLS counts ($C_{dls,i}$), and promiscuous binding counts ($C_{promiscuity, i}$).  Hence, 
\begin{equation}
\mu_{i,NTC} = \sigma(B_{NTC,i} + \beta'_{DLS} p_{dls,i}+\beta'_{promiscuity}c_ {promiscuity} +\beta'_{constant}).
\end{equation}

We examine two different ways of estimating the proportion of product $k$, $p_{k}$. The first is to set these values based on our internal reaction yield dataset: $p_{k, lab}$. To refine this approach, and take into account the possible noise in the yield measurements, we propose to adjust the estimated proportion with a learned parameter for each molecule: $p_{k, adjust}$. In this case, $p_{k} = \sigma(p_{k, adjust} + p_{k, lab})$.

To learn the enrichment values, $R_{target, molecule}$ and $R_{NTC,molecule}$, we employ  two similar models. The input to both is a molecular graph corresponding to a chemical product. The Weave atom featurizer \cite{Kearnes_2016} is employed to encode the atoms and a canonical bond featurizer \cite{dgllife} is used to encode the bonds between atoms. For the first model, we use a single message passing graph neural network \cite{dgllife} followed by a sequence to sequence layer \cite{vinyals2015order}. This outputs a 128 dimensional encoding that is transformed using two distinct fully connected networks into $R_{target, i}$ and $R_{NTC, i}$. The second model is identical except that the 128 dimensional encoding is also transformed into $p_{k, adjust}$ through an additional fully connected network. These models are tuned on the training set using batch sizes of 32 and an Adam optimizer \cite{Adam} with a learning rate of $10^{-3}$ for 15 epochs. 

Following \cite{ma_regression_2021}, for a given DNA barcode example $i$, the negative binomial loss for the target and NTC values can be written as
\begin{equation}
L_{value,i} = -\log(P(c_{value,i} \mid \mu_{value, i}, \alpha_{value})
\end{equation}
where the $P$ is the probability mass function of the negative binomial distribution  parameterized by $\mu_{value, i}$ and $\alpha_{value}$. The full loss for example $i$
is 
\begin{equation}
L_i =  L_{target,i} + L_{ntc,i} + \gamma\cdot (R_{target,i}^2 + R_{NTC,i}^2 )
\end{equation}

During the validation phase of the model when it is used to screen external models, the $R_{target, i}$ values are used to rank the binding affinities of the compounds to the target. 

\section{Results and Discussion}

We evaluate our model with the full product proportion data against three different models. The first is a model where the only source of binding enrichment is the trisynthons. ($B_{target, i}=p_{tri, i}R_{tri, i}$, $B_{NTC, i}=p_{tri, i}R_{NTC,tri, i}$). In the second model, the only source of binding enrichment is the disynthons. ($B_{target, i}= \Sigma_{j=1}^{3}p_{di_j,i}R_{di_j,i}$, $B_{NTC,i}= \Sigma_{j=1}^{3}p_{di_j,i}R_{NTC,di_j,i}$). 
\begin{figure}[h]
\begin{center}
\includegraphics[scale = 0.25]{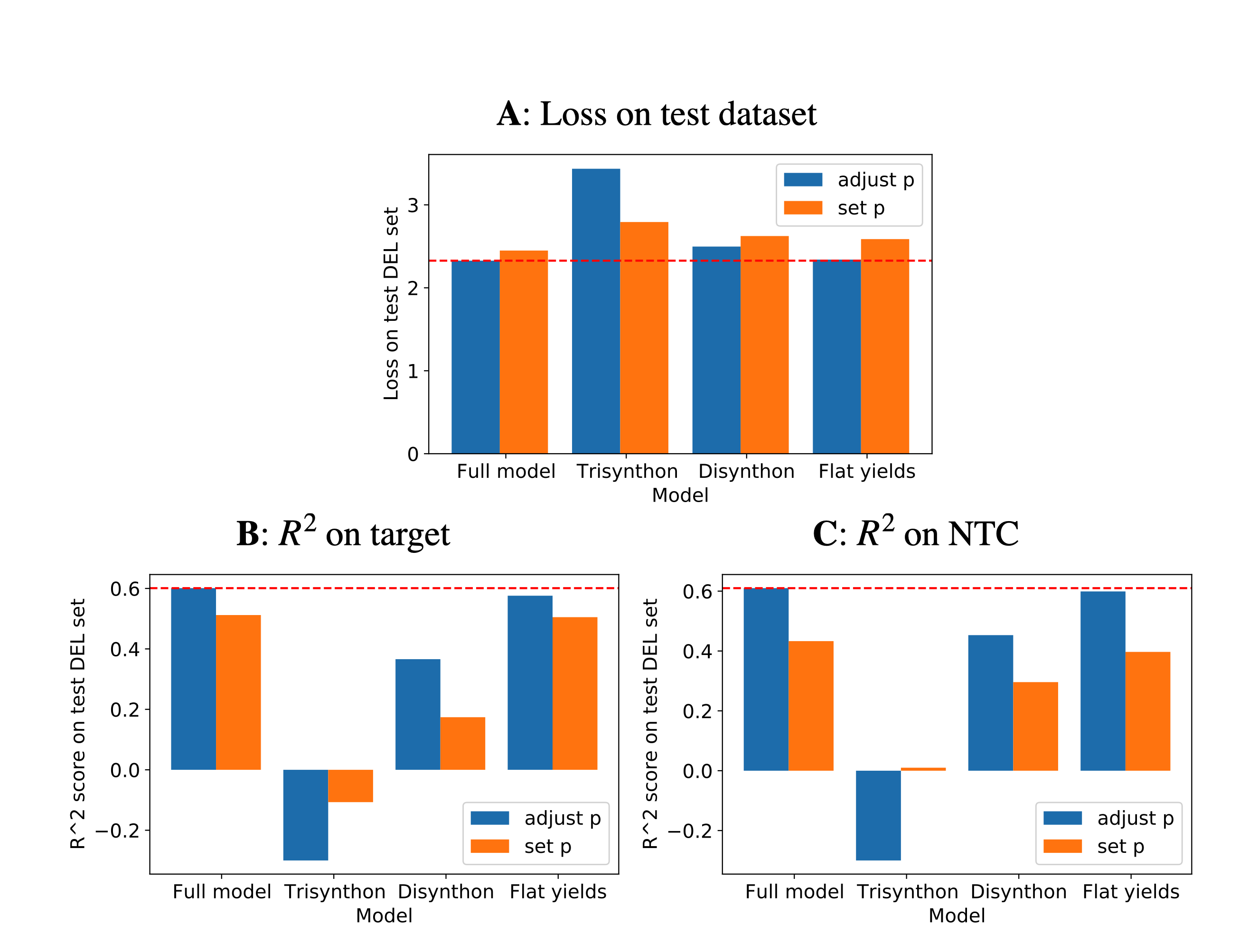} 
\end{center}
\caption{Loss and $R^2$ scores on the test data set. Figure A shows the negative binomial loss on the test data set. Figure B shows the the $R^2$ score between the target protein counts and the generated $\mu_{target,i}$ counts. Figure C shows the the $R^2$ score between the NTC counts and the generated $\mu_{NTC,i}$ counts.}
\label{delres}
\end{figure}

In the third model, we assume that each reaction produces a $70\%$ yield, so all $p_{di_j,i} = 0.147$ and all $p_{tri, i} = 0.147$. For each of the four models described above, we ran an experiment where the product proportions are fixed and an additional experiment where an adjustment to the proportions is learned as described above. 
The results are shown in figure  \ref{delres}.
\begin{figure}[h]
\begin{center}
\includegraphics[scale = 0.4]{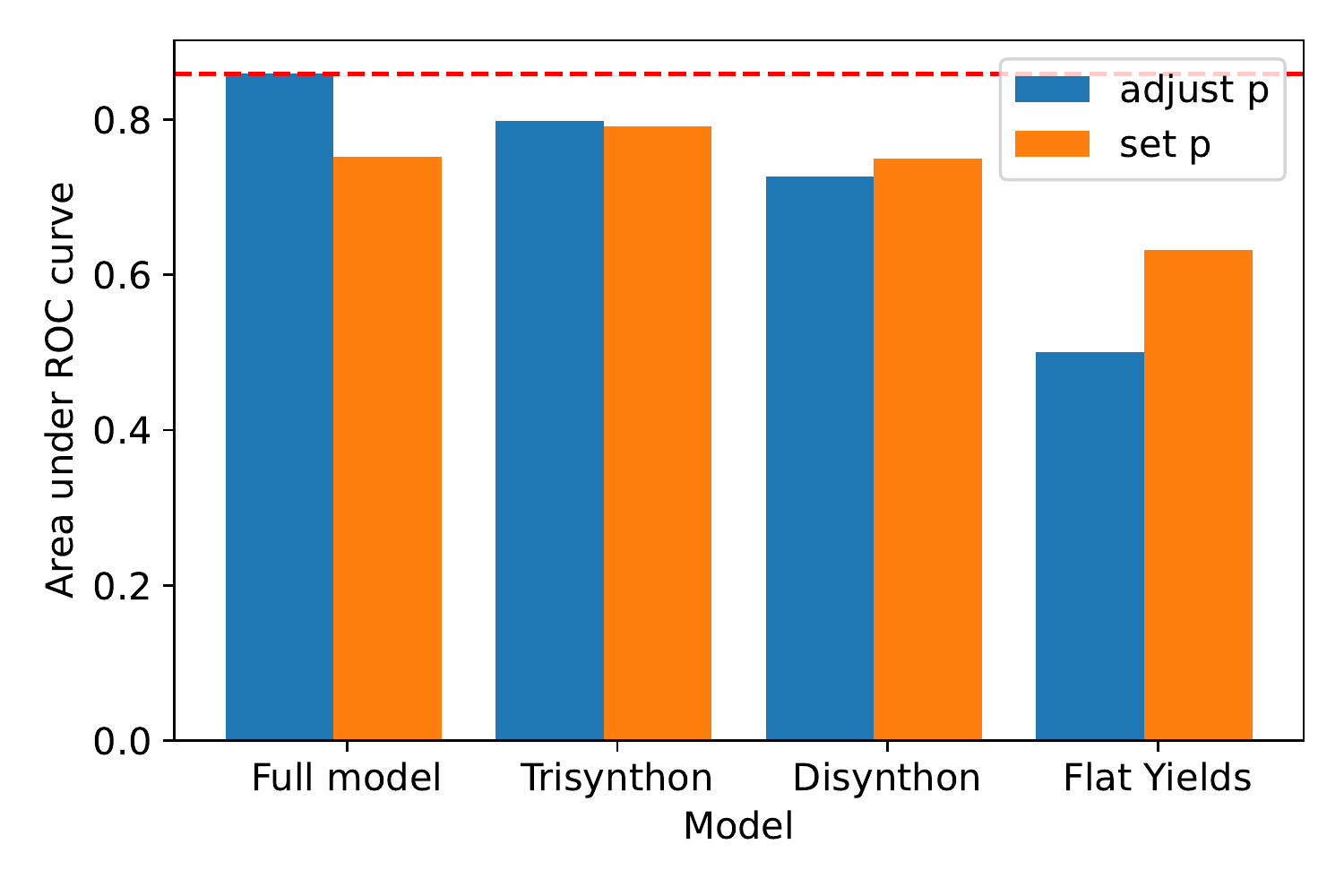}

\end{center}
\caption{Area under the ROC curve on 150 external molecules for various models}
\label{ROC}
\end{figure}

One evaluation metric is performance on the testing set which is a subset of the internal DEL data. When we evaluate the model against testing data from the internal DEL data set, we calculate the loss and examine the $R^2$ scores of between the $\mu_{ntc}$ and $\mu_{target}$ values and the true counts.

Across the two approaches to modelling the product proportions, the trisynthon-only model produces noticeably worse results. This is likely evidence of the fact that the trisynthons data by itself is too noisy to produce reliable estimates and does not capture all of the DEL products. The disynthon-only model produces  worse results than the full model but better results than the trisynthon model. This suggests that a large amount of the binding of a DNA tag can be explained by the disynthon products. It appears that modeling disynthons is not only a de-noising step, but is also an integral portion of the data. The model with constant yields performs slightly worse, highlighting the importance of modeling the yield and product proportion values. The model with the learned adjustments to yields shows improved performance, likely due to noise in the laboratory processes where several rounds of purification are performed. 

To estimate the generalization ability of the learned model, we test its performance on a dataset of 150 molecules from commercial vendors. The binding affinities were measured internally, and the molecules were classified as binders or non-binders to the target. The enrichment values $R_{target, moecule}$ are used for predicting the binding affinity. The area under the receiver operating characteristic (ROC) curve for various models is shown in figure \ref{ROC}. On this dataset, the full model with learned yields outperforms the remaining models. 

\begin{figure}[h]
\begin{center}
\includegraphics[scale = 0.4]{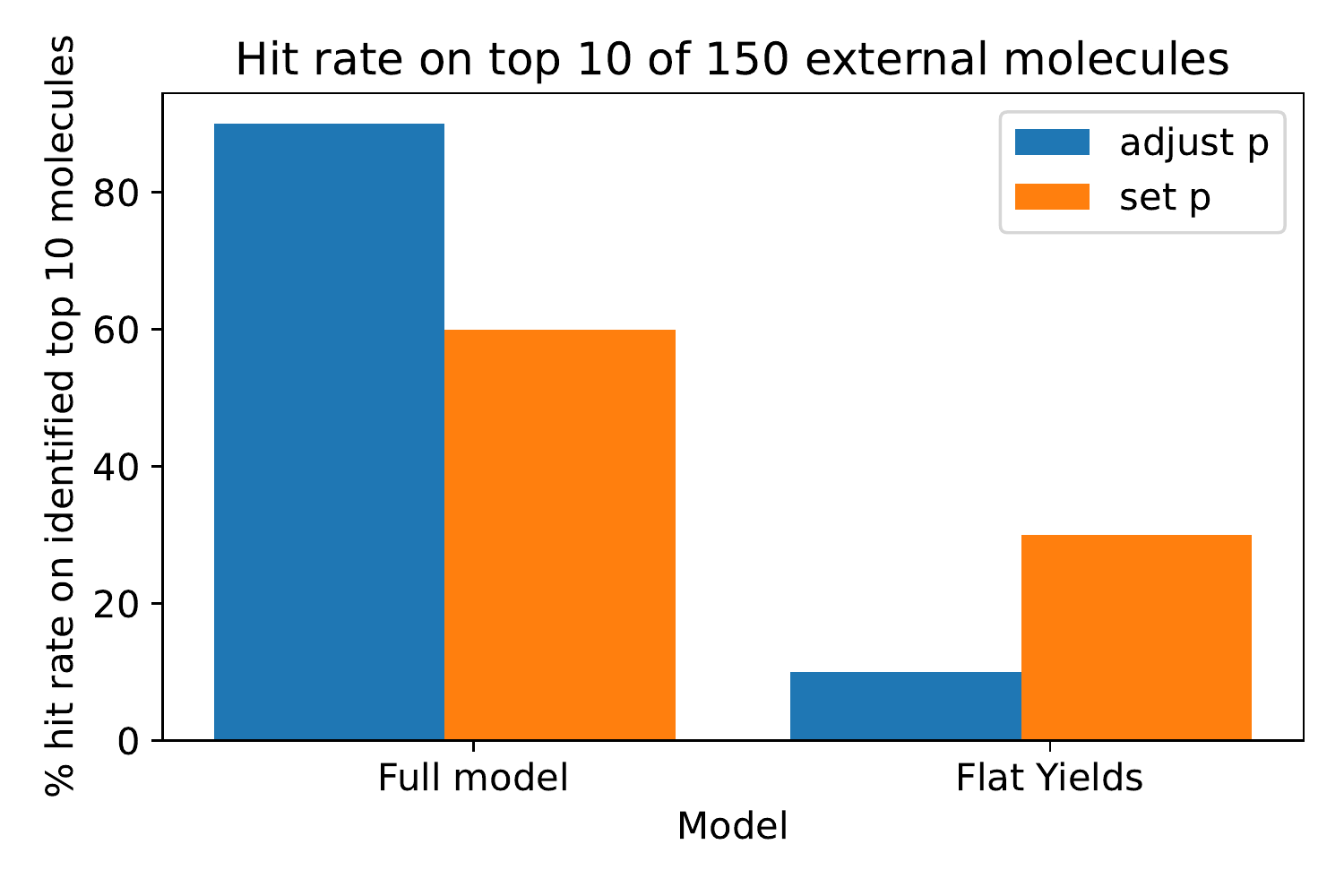}

\end{center}
\caption{Hit rate on top 10 of 150 external molecules}
\label{HitRate}
\end{figure}

To further investigate the effects of incorporating yield data, we evaluated the full model and the model with flat yields on the 150 external molecules. We examined the hit rate percent among the top 10 identified hits, and the results are shown in figure \ref{HitRate}. This demonstrates that the model with lab yields and a learned adjustment performs best in a virtual screening application and that setting learned yields makes more of an impact in pertinent individual cases. 
\section{Conclusion}
    
DNA encoded libraries are powerful tools for screening large small molecule libraries against a target protein. In this work, we present a modeling approach that leverages information about full and partial products in DELs and their proportions. We demonstrate that our approach that models full and partial products as well as their proportions outperforms models with only some of the products or a model that does not include proportions of the products. Using only full product (trisynthon) data can produce datasets that are too noisy to support effective training, while also inadequately describing the underlying data. Using only partial (disynthon) data is an effective approach to aggregating and de-noising DEL data, but altogether disregards potentially useful data.  Moreover, our approach can be used to identify molecules on an external data set that are strong binders to a target protein of interest. 

\bibliography{DEL}
\bibliographystyle{iclr2022_conference}

\end{document}